\documentclass[11pt]{article}

\usepackage[final]{acl}

\usepackage{times}
\usepackage{latexsym}

\usepackage[T1]{fontenc}

\usepackage[utf8]{inputenc}

\usepackage{microtype}

\usepackage{inconsolata}

\usepackage{graphicx}
\usepackage{amsmath}
\usepackage{booktabs}
\usepackage{multirow}
\usepackage{makecell}
\usepackage[table]{xcolor}

\definecolor{heatmax}{HTML}{FDBE85}
\definecolor{heatmin}{HTML}{FFF2CC}
\definecolor{BrickRed}{RGB}{203,65,84}
\definecolor{ForestGreen}{RGB}{34,139,34}

\usepackage{siunitx}
\usepackage{dblfloatfix}

\title{Test-Time Verification for Text-to-SQL via Outcome Reward Models}

\author{
\textbf{Mattia Tritto\textsuperscript{1}},
\textbf{Giuseppe Farano\textsuperscript{1}},
\textbf{Dario Di Palma\textsuperscript{1}},
\textbf{Gaetano Rossiello\textsuperscript{2}},
\\
\textbf{Fedelucio Narducci\textsuperscript{1}},
\textbf{Dharmashankar Subramanian\textsuperscript{2}},
\textbf{Tommaso Di Noia\textsuperscript{1}}
\\
\\
\textsuperscript{1}Polytechnic University of Bari, Bari, Italy\\
\textsuperscript{2}IBM T.J. Watson Research Center, Yorktown Heights, NY, USA
}

\begin{document}
\maketitle
\begin{abstract}
Improving the reliability of large language models (LLMs) at inference time is a central challenge in structured reasoning tasks such as Text-to-SQL. Common test-time inference strategies, including Best-of-$N$ sampling and Majority Voting, rely on heuristic signals such as execution success or output frequency, which provide limited semantic discrimination across candidate outputs.
In this work, we study \emph{Outcome Reward Models} (ORMs) as learned semantic scoring functions for test-time verification in Text-to-SQL. While ORMs have been previously explored for test-time scaling and alignment, their application to structured query generation remains underexplored. We introduce \textit{GradeSQL}, a scalable framework for training task-specific ORMs via automated candidate generation and execution-based labeling, enabling verifier training without manual annotation.
We integrate ORMs into a verification-driven Best-of-$N$ pipeline and evaluate our approach on the BIRD and Spider benchmarks across multiple open-source LLM families. ORM-based selection consistently outperforms execution-based Best-of-$N$ and Majority Voting, with gains of up to +4.33\% on BIRD and +2.10\% on Spider. We further show that ORMs scale effectively with larger candidate sets and yield stronger improvements on complex queries.
Overall, our results demonstrate that ORM-based verification provides a simple, effective, and scalable alternative to heuristic test-time selection strategies for Text-to-SQL. 
Code\footnote{\href{https://github.com/sisinflab/GradeSQL}
{GradeSQL Framework}}, 
datasets\footnote{
\href{https://huggingface.co/collections/sisinflab-ai/gradesql-training-datasets-68ac62e1356b5399ef81236c}{GradeSQL Training Datasets}}, 
and models\footnote{\href{https://huggingface.co/collections/sisinflab-ai/gradesql-models-68ac58755ffe5fe880e0acb5}{Pretrained GradeSQL ORMs}} are publicly available.
\end{abstract}

\section{Introduction}

As large language models (LLMs) are increasingly deployed in structured reasoning tasks, improving reliability at inference time has become a central challenge. While scaling model size and training data has driven substantial progress, further gains increasingly depend on \emph{post-training} techniques that leverage verification, feedback, and candidate selection mechanisms. In this setting, test-time inference plays a critical role: rather than producing a single output, models generate multiple candidates and rely on selection strategies to identify the most accurate solution.

Text-to-SQL, the task of translating natural language questions into executable SQL queries, provides a natural testbed for studying test-time verification. It enables intuitive access to structured databases~\cite{DBLP:conf/iceis/NascimentoGFVIO24,DBLP:journals/pvldb/KimSHL20}, but requires strict semantic correctness: small errors in query structure can lead to incorrect results. Despite strong progress with LLMs~\cite{DBLP:journals/air/Kumar24,DBLP:conf/nips/PourrezaR23}, performance remains limited on complex queries involving multi-table joins, nested subqueries, and subtle constraints~\cite{DBLP:journals/corr/abs-2407-19517}. This makes robust verification mechanisms essential.

A common approach to improving performance is to increase test-time compute via strategies such as Best-of-$N$ (BoN) sampling and Majority Voting~\cite{DBLP:journals/corr/abs-2110-14168,DBLP:journals/corr/abs-2505-13271}. These methods exploit generation diversity, but rely on heuristic signals such as execution success or output frequency, which provide only coarse proxies for semantic correctness. As candidate sets grow, these heuristics often saturate and fail to discriminate between semantically valid and invalid queries.

Reward models offer a principled alternative by learning to score outputs according to task-specific correctness. In particular, \emph{Outcome Reward Models} (ORMs) assign scalar scores to complete outputs and have been successfully used for test-time scaling and alignment in prior work~\cite{DBLP:journals/corr/abs-2110-14168}. However, their application to structured reasoning tasks such as Text-to-SQL remains underexplored, especially in terms of scalable data curation, task-specific training, and integration into practical inference pipelines.

In this work, we investigate how ORMs can be effectively adapted and deployed as verifiers for Text-to-SQL. Our goal is not to introduce a new class of reward models, but to demonstrate how existing ORM paradigms can be trained and applied in a scalable way for structured query generation. A key challenge is data availability: training ORMs requires labeled candidate outputs, which are scarce for structured tasks.

To address this, we introduce \textit{GradeSQL}, a framework for scalable ORM training based on automated data synthesis. Given a natural language question and database schema, we generate diverse candidate SQL queries, label them via execution equivalence, and fine-tune a verifier to predict semantic correctness. At inference time, the trained ORM is used to re-rank candidates within a Best-of-$N$ pipeline, replacing heuristic selection with learned scoring.

We evaluate our approach on the BIRD and Spider benchmarks using multiple open-source LLM families. Results show that ORM-based verification consistently improves execution accuracy over heuristic baselines, with gains of up to +4.33\% on BIRD and +2.10\% on Spider. While absolute improvements are moderate, they are consistent across models and datasets, and are obtained without modifying the generator or requiring additional environment interaction at inference time. We further show that ORM-based selection scales more effectively with larger candidate budgets and yields stronger gains on complex queries.

In summary, our contributions are:
\begin{itemize}
\item \textbf{ORM-based Verification for Text-to-SQL:} We study Outcome Reward Models as learned semantic scorers for test-time candidate selection in structured query generation.
\item \textbf{Scalable Data Curation Pipeline:} We propose an automated framework for generating and labeling candidate SQL queries, enabling ORM training without manual annotation.
\item \textbf{Empirical Analysis of Test-Time Verification:} We provide a systematic evaluation of ORM-based selection across models, datasets, and inference regimes, showing consistent improvements over heuristic baselines.
\end{itemize}

\section{Related Work}\label{sec:related}

\subsection{Text-to-SQL with LLMs}
Text-to-SQL translates natural language questions into executable SQL queries, enabling natural language access to structured databases. The field has evolved from rule-based and logic-driven systems~\citep{woods1972lunar,DBLP:journals/coling/WarrenP82} to neural semantic parsing approaches~\citep{DBLP:conf/emnlp/TangM00,DBLP:conf/acl/XiaoDG16}. The introduction of large-scale benchmarks such as Spider~\citep{DBLP:conf/emnlp/YuZYYWLMLYRZR18} enabled the development of schema-aware models based on pre-trained language models (PLMs) such as BERT and T5~\citep{DBLP:conf/naacl/DevlinCLT19,DBLP:journals/jmlr/RaffelSRLNMZLL20}, with architectures like RAT-SQL highlighting the importance of schema encoding~\citep{DBLP:conf/acl/WangSLPR20}.

Recent advances with large language models (LLMs) have further improved performance through zero- and few-shot prompting, task decomposition, and retrieval-augmented generation~\citep{DBLP:journals/air/Kumar24,DBLP:conf/nips/PourrezaR23,DBLP:conf/nips/LiHQYLLWQGHZ0LC23}. Despite these advances, Text-to-SQL remains challenging for complex queries involving multi-step reasoning and strict semantic correctness, motivating stronger verification mechanisms at inference time.

\subsection{Test-Time Inference and Verification}
Test-time inference (TTI) improves model outputs by allocating additional computation at inference without modifying model parameters~\citep{DBLP:conf/iclr/Snell0XK25}. A common approach is to generate multiple candidate outputs and select among them. Best-of-$N$ sampling and related strategies have been widely studied, including in mathematical reasoning where learned verifiers are used to select correct solutions from multiple candidates~\citep{DBLP:journals/corr/abs-2110-14168}. Majority Voting further aggregates multiple outputs based on answer consistency.

While these approaches are effective, many practical implementations rely on heuristic signals such as execution success, output agreement, or frequency, which may fail to capture semantic correctness in structured tasks. This has motivated the use of learned verification models that score candidate outputs based on task-specific criteria.

\begin{figure*}[t]
    \centering
    \includegraphics[width=0.95\textwidth]{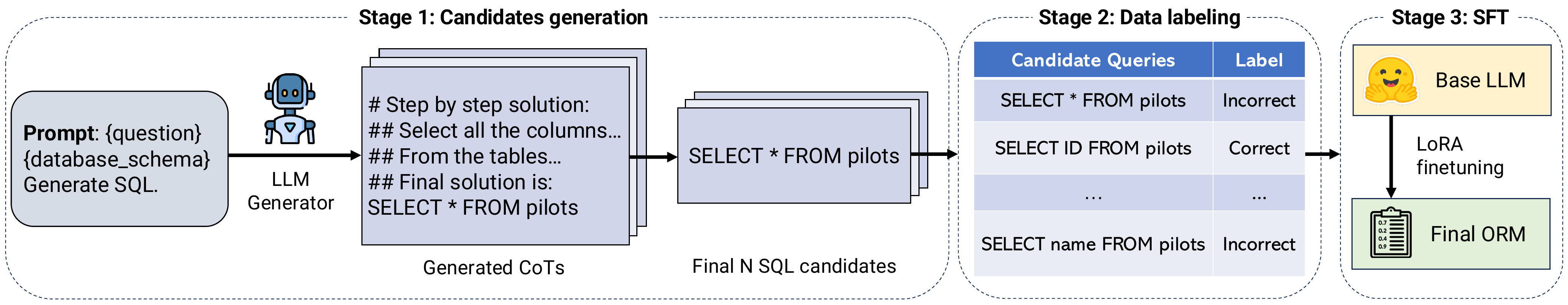}
    \caption{Overview of the \textit{GradeSQL} framework for training an ORM. The framework consists of three stages: (i) \textbf{Candidate Generation}, (ii) \textbf{Data Labeling}, and (iii) \textbf{Supervised Fine-Tuning (SFT)}.}
    \label{fig:orm_framework}
\end{figure*}

\subsection{Reward Models and Learned Verifiers}
Reward models provide a principled framework for scoring model outputs. Process Reward Models (PRMs) evaluate intermediate reasoning steps, while Outcome Reward Models (ORMs) assign scalar scores to final outputs~\citep{DBLP:journals/corr/abs-2110-14168}. ORMs have been successfully used for test-time scaling and alignment, particularly in reasoning tasks where multiple candidate solutions can be generated and ranked.
More recent work explores generative verifiers~\citep{DBLP:conf/iclr/ZhangHBKKA25}, where verification is framed as next-token prediction rather than explicit reward modeling, enabling scalable training without separate reward heads. These approaches further highlight the flexibility of learned verification mechanisms in post-training pipelines.

In Text-to-SQL, learned verification remains relatively underexplored. Prior work has focused primarily on improving generation through prompt design, retrieval, or fine-tuning, rather than explicitly training verifiers. This leaves open the question of how to effectively train and deploy task-specific reward models for structured query verification.

\subsection{Data Curation for Verifier Training}
A key challenge in training reward models is the availability of labeled data. Existing Text-to-SQL datasets are designed for generation rather than verification, typically providing a single gold query per input. Synthetic data generation approaches have been proposed to scale training data~\citep{DBLP:journals/pvldb/LiWZHZJWZCSCL25,DBLP:conf/acl/YangHY0LZ24}, but are not tailored to the needs of verifier training, which requires diverse candidate outputs and fine-grained correctness signals.

In this work, we address this gap by introducing a scalable pipeline for generating and labeling candidate SQL queries, enabling the training of task-specific Outcome Reward Models for test-time verification.


\section{Methodology}\label{sec:methods}

\subsection{Problem Formulation}
Given a natural language question $q_i \in Q$ and its associated database schema $\Sigma_i$, the Text-to-SQL task aims to generate a correct SQL query $y_{gold_i} \in Y_{\text{gold}}$~\citep{DBLP:conf/acl/BaekSHSSGB25, DBLP:conf/acl/DaiYMC25}. Under a test-time inference setting, the goal extends from producing a single query to selecting the best query from a set of candidates generated by a large language model (LLM).

Specifically, for a given question $q_i$, an LLM produces a candidate set
\[
Y_{\text{candidate}} = \{c_1, c_2, \dots, c_N\},
\]
from which a verifier selects the most semantically faithful query. In this work, the verifier is an \emph{Outcome Reward Model (ORM)}, formalized as a scoring function
\[
\phi(q_i, c_j) \in [0,1],
\]
which estimates the likelihood that candidate $c_j$ correctly captures the intent of $q_i$. The final prediction is obtained via:
\[
c^* = \operatorname*{arg\,max}_{c_j \in Y_{\text{candidate}}} \phi(q_i, c_j).
\]

Training such a verifier poses two challenges: (i) constructing datasets with labeled candidate correctness, and (ii) assigning scores that reflect semantic alignment rather than surface heuristics. To address this, we leverage the LLM’s self-evaluation capabilities~\citep{DBLP:conf/naacl/LiuLZDCHXCW24, DBLP:conf/emnlp/HuangQLYXZL24, DBLP:conf/emnlp/LiWFZWC24}, prompting it to assess candidate correctness and using the resulting logits as supervision signals.

We implement this approach in the \textbf{GradeSQL} framework, illustrated in Figure~\ref{fig:orm_framework}.

\begin{figure}[t]
    \centering
    \includegraphics[width=0.45\textwidth]{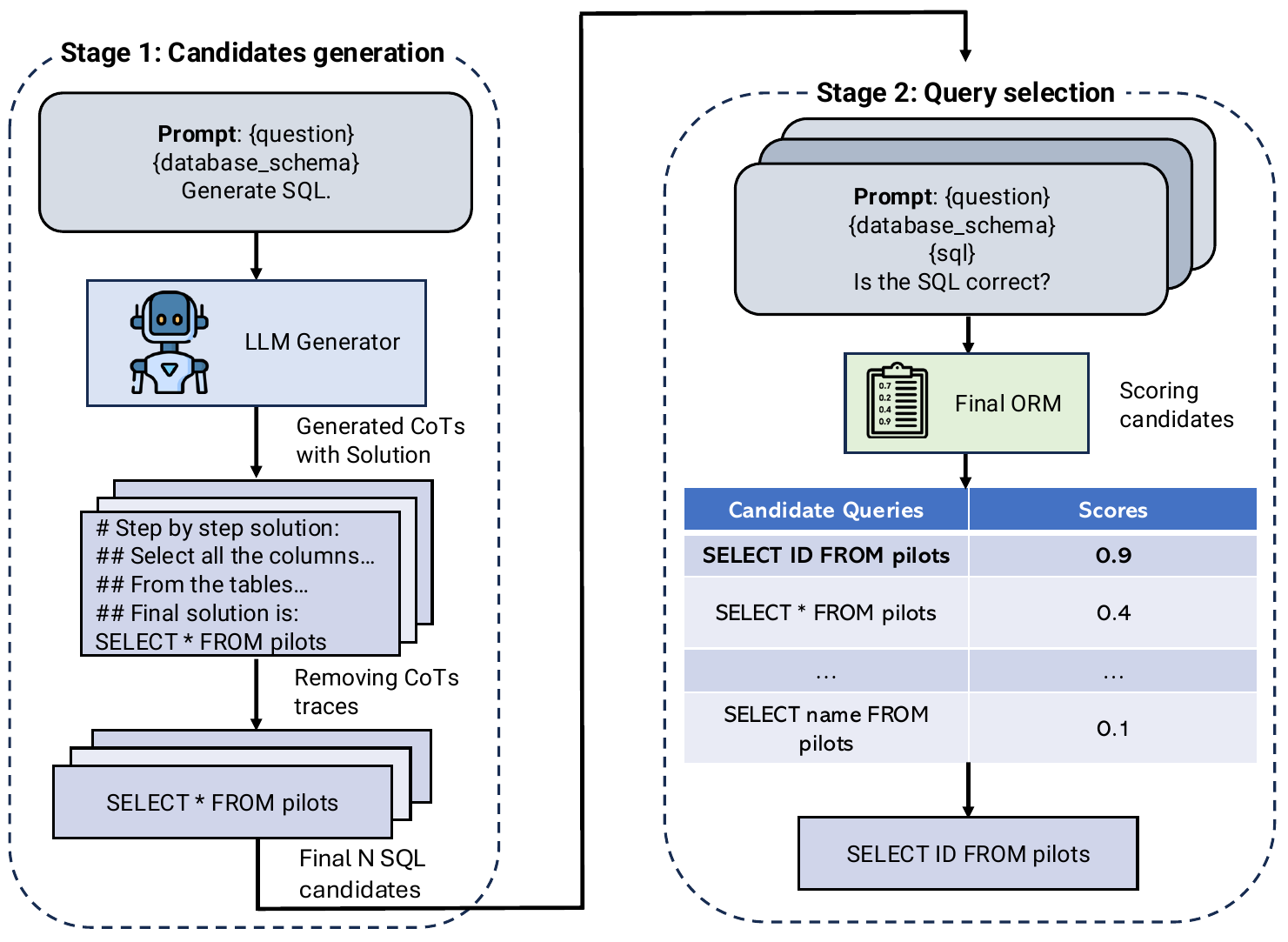}
    \caption{ORM-based inference pipeline: candidate SQL queries are generated by an LLM and ranked by the ORM, with the highest-scoring query selected.}
    \label{fig:inference_schema}
\end{figure}

\subsection{GradeSQL Framework}
Given a dataset $\mathcal{D} = \{(q_i, y_{gold_i})\}_{i=1}^n$, GradeSQL trains an ORM through three stages.

\noindent\textbf{Stage 1: Candidate Generation}  
For each question $q_i$ and schema $\Sigma_i$, a generator LLM $\mathcal{G}$ produces a set of $N$ candidate SQL queries:
\[
Y_{\text{candidate}} = \mathcal{G}(q_i, \Sigma_i) = \{c_1, \dots, c_N\}.
\]
To promote diversity in the candidate set, generation is performed using stochastic decoding strategies, allowing the model to explore multiple plausible query formulations. This diversity is essential for effective verification, as it exposes the ORM to a broader range of correct and incorrect candidates, including semantically equivalent queries with different syntactic structures as well as near-miss errors.

\noindent\textbf{Stage 2: Data Labeling}  
Each candidate is executed on the database. Let $R(c_j)$ denote the result set of query $c_j$. A candidate is labeled as \textit{correct} if $R(c_j) = R(y_{gold_i})$ and \textit{incorrect} otherwise. Queries that raise execution errors are discarded. The labeling function is:
\[
\ell(c_j) =
\begin{cases}
\text{Yes} & \text{if } R(c_j) = R(y_{gold_i}), \\
\text{No} & \text{if } R(c_j) \neq R(y_{gold_i}), \\
\text{discarded} & \text{if execution fails}.
\end{cases}
\]

\noindent\textbf{Stage 3: Supervised Fine-Tuning}  
The labeled dataset $\mathcal{D}_{\text{label}} = \{(c_j, \ell(c_j))\}$ is used to fine-tune a verifier LLM as a binary autoregressive classifier. Each input is constructed as
\[
x = \mathrm{Prompt}(\Sigma, q, c),
\]
with target label $l \in \{\texttt{Yes}, \texttt{No}\}$. We fine-tune the model using LoRA~\citep{DBLP:conf/iclr/HuSWALWWC22} under the causal language modeling objective~\citep{radford2019language}:
\[
\mathcal{L}(\theta) = - \sum_{t=1}^{|s|} \log P_\theta(s_t \mid s_{<t}),
\]
where $s = [x; l]$.

\subsection{ORM Inference and Probabilistic Scoring}
At inference time, the ORM assigns a probabilistic score to each candidate SQL query. Given a prompt $x = \mathrm{Prompt}(\Sigma, q, c)$, the verifier defines:
\[
P(y \mid x; \theta), \quad y \in \{\texttt{Yes}, \texttt{No}\},
\]
and uses the probability of the ``\texttt{Yes}'' token as the candidate score:
\[
\text{Score}(c) = P(y_{\text{yes}} \mid x; \theta).
\]

Rather than making a hard classification decision, this formulation yields a continuous confidence score that enables robust ranking of candidates (Figure~\ref{fig:inference_schema}). Probabilistic scoring offers fine-grained calibration across syntactically valid queries and avoids the brittleness of binary thresholding, making it well-suited for test-time verification.

\paragraph{Interpretation as a Learned Verifier}
The ORM can be interpreted as a learned verifier that approximates semantic correctness beyond simple execution-based heuristics. While execution signals provide a binary notion of correctness, the ORM learns a softer scoring function that captures patterns across candidate queries, such as structural consistency, alignment with schema elements, and robustness to spurious matches. In particular, by training on diverse candidate sets that include both correct and incorrect queries, the ORM is exposed to fine-grained distinctions between semantically valid and invalid formulations. This allows it to assign higher confidence to candidates that are more likely to generalize beyond execution equivalence alone, especially in cases where multiple queries execute successfully but differ in semantic fidelity. As a result, the ORM complements heuristic selection by providing a more discriminative ranking signal over the candidate pool.

\section{Experimental Setup}\label{sec:exp_setup}

\textbf{Datasets}  
We evaluate on two cross-domain Text-to-SQL benchmarks: Spider~\citep{DBLP:conf/emnlp/YuZYYWLMLYRZR18} and BIRD~\citep{DBLP:conf/nips/LiHQYLLWQGHZ0LC23}. Both use disjoint databases across splits. For Spider, we train ORMs on the training set and evaluate on dev and test. For BIRD, we train on the training set and evaluate on the dev set due to the hidden test split.

\noindent\textbf{Metrics}  
We use Execution Accuracy (EX), which measures whether predicted and gold queries return identical results, and Pass@$N$, which measures whether at least one of the $N$ generated candidates is correct.

\noindent\textbf{Baselines}  
We compare against two test-time inference strategies: Majority Voting and execution-based Best-of-$N$, both selecting a single query from a candidate pool without using gold references.

\paragraph{Majority Voting}  
Candidates are executed and grouped by identical result sets. The largest group is selected and a query is sampled uniformly from it~\citep{DBLP:journals/corr/abs-2505-13271}.

\paragraph{Execution-based Best-of-$N$}  
Candidates are ranked using a simple execution heuristic that favors queries that execute successfully and return non-empty results~\citep{DBLP:conf/iclr/ChenZNZLLC23}.

\begin{table}[t!]
\centering
\caption{Reproducibility results of OmniSQL-7B. Maj = Majority Voting (temperature $T{=}0.8$, $N{=}8$); Gre = Greedy decoding ($N{=}1$). The table compares our reproduction (Our) with the original results (Origin) on the BIRD dev, Spider dev, and Spider test benchmarks.}
\label{tab:omnisql_reproducibility}
\renewcommand{\arraystretch}{1.08}
\setlength{\tabcolsep}{6pt}

\resizebox{0.95\columnwidth}{!}{%
    \begin{tabular}{lcccccc}
    \toprule
     & \multicolumn{2}{c}{\textbf{BIRD dev}} & \multicolumn{2}{c}{\textbf{Spider dev}} & \multicolumn{2}{c}{\textbf{Spider test}} \\
    \cmidrule(lr){2-3} \cmidrule(lr){4-5} \cmidrule(lr){6-7}
     & Maj & Gre & Maj & Gre & Maj & Gre \\
    \midrule
    \textbf{Our}    & 66.95 & 64.41 & 83.95 & 82.11 & 85.61 & 84.49 \\
    \textbf{Origin} & 66.10 & 63.90 & 81.60 & 81.20 & 88.90 & 87.90 \\
    \bottomrule
    \end{tabular}
}

\end{table}

\noindent\textbf{ORMs and Evaluation Setup}  
We use OmniSQL-7B~\citep{DBLP:journals/pvldb/LiWZHZJWZCSCL25} as the generator. For each question, the same set of $N$ candidates is produced and shared across all methods. We compare Majority Voting, execution-based Best-of-$N$, and ORM-based Best-of-$N$ over this fixed pool.

This setup isolates the effect of the selection mechanism from generation quality. All methods operate on identical candidates and are evaluated under the same metric. ORM-based selection requires an additional offline training phase, so results reflect differences in selection quality rather than fully compute-matched end-to-end systems.

To validate the generator, we reproduce OmniSQL results on Spider and BIRD. Table~\ref{tab:omnisql_reproducibility} shows deviations within 1\% on BIRD dev and within 3\% on Spider dev.

\noindent\textbf{Compute Considerations}  
Heuristic baselines require no additional training, whereas ORM-based selection incurs offline verifier training. At inference time, all methods operate on the same candidate pool. The trade-off is between upfront training cost and improved semantic discrimination during selection. In our setting, training is amortized across queries and requires no additional environment interaction beyond candidate generation and execution.

\noindent\textbf{Reproducibility Details}  
Experiments were run on nodes with Intel Xeon 8358 CPUs, 512 GB RAM, and 4 NVIDIA A100 GPUs (64 GB). Candidates were generated using \texttt{vLLM}, and ORM inference used Hugging Face \texttt{transformers}. All runs use seed 42.


\section{Results and Discussion}\label{sec:discus}

\begin{table}[t]
\centering
\caption{Execution accuracy (\%) of OmniSQL-7B on \textit{BIRD dev}, \textit{Spider dev}, and \textit{Spider test}. 
$\Delta$ indicates the gain over the baseline (\emph{$N{=}1$}). 
Best and runner-up results are in \textbf{bold} and \underline{underlined}, respectively. 
Values marked with $\dagger$ are not McNemar significant ($p<0.05$), and those with * lose significance after Bonferroni correction.}

\label{tab:main_results}



\resizebox{0.95\columnwidth}{!}{%
  \setlength{\tabcolsep}{5pt}
  \begin{tabular}{lcccccc}
    \toprule
    \multirow{2}{*}{\textbf{Method}} &
      \multicolumn{2}{c}{\textbf{BIRD dev}} &
      \multicolumn{2}{c}{\textbf{Spider dev}} &
      \multicolumn{2}{c}{\textbf{Spider test}} \\
    \cmidrule(lr){2-3}\cmidrule(lr){4-5}\cmidrule(lr){6-7}
    & EX & $\Delta$ & EX & $\Delta$ & EX & $\Delta$ \\
    \midrule
    Baseline (N=1) & 63.89 & -- & 82.40 & -- & 84.02 & -- \\
    Majority Voting (N=32)  & \underline{66.95} & +3.06 & \underline{83.75}$*$ & +1.35 & \underline{85.47} & +1.45 \\
    \multicolumn{7}{l}{Best-of-$N$ (N=32)} \\
    \quad Execution-based & 66.04 & +2.15 & 82.79$\dagger*$ & +0.39 & 85.14 & +1.12 \\
    \quad ORM-based & \textbf{68.90} & +5.01 & \textbf{84.53}$*$ & +2.13 & \textbf{87.47} & +3.45 \\
    \bottomrule
  \end{tabular}%
}
\end{table}

We evaluate our framework across seven open-source LLMs: Qwen2.5-1.5B/7B-Instruct, Granite-3.3-2B/8B-Instruct, Llama-3.2-1B, Llama-3.1-8B-Instruct, and OmniSQL-7B. Experiments are conducted on BIRD dev (1,534 samples), Spider dev (1,034 samples), and Spider test (2,147 samples). ORM training data is constructed from the original training splits (9,428 queries for BIRD and 9,000 for Spider), yielding 9,411 and 8,960 queries after light preprocessing.

For each question, OmniSQL-7B generates $N=32$ SQL candidates, following prior work~\citep{DBLP:conf/acl/LiL025} as a trade-off between performance and computational cost~\citep{DBLP:conf/iclr/0002WSLCNCZ23,DBLP:conf/nips/LewkowyczADDMRS22}. Invalid queries are discarded, and remaining candidates are labeled based on execution equivalence with the gold query, resulting in 82,640 labeled samples for BIRD and 50,073 for Spider.

\noindent\textbf{ORM vs. Test-Time Baselines}
Table~\ref{tab:main_results} compares ORM-based Best-of-$N$ with execution-based Best-of-$N$ (ex-BoN) and Majority Voting (Maj), using OmniSQL-7B as the generator. Statistical significance is assessed using the McNemar test with Bonferroni correction~\citep{DBLP:conf/acl/ReichartDBS18}. ORM-based selection achieves the highest execution accuracy across all datasets. Except for Spider dev (likely due to smaller sample size), improvements remain statistically significant.

On BIRD dev, ORM reaches 68.90\%, outperforming Maj (66.95\%) and ex-BoN (66.04\%), with a +5.01 gain over the $N{=}1$ baseline. On Spider dev, ORM achieves 84.53\% (+2.13 over Maj), and on Spider test 87.47\%, exceeding both Maj (85.47\%) and ex-BoN (85.14\%) by a wide margin.

\begin{table}[t]
\centering
\caption{Dataset statistics for \textit{BIRD} and \textit{Spider} (training splits). We report (i) the number of questions before and after de-duplication (with removals $\Delta$), and (ii) the final size of the labeled sets generated by sampling $N\!=\!32$ candidates per question. Percentages indicate the class distribution \emph{(Incorrect / Correct)} in the \emph{imbalanced} set; the balanced set is 50/50.}
\label{tab:dataset_stats}

\setlength{\tabcolsep}{6pt}

\resizebox{0.95\columnwidth}{!}{%
\begin{tabular}{@{}lrr@{}}
  \toprule
  & \textbf{BIRD (train)} & \textbf{Spider (train)} \\
  \midrule
  \multicolumn{3}{@{}l}{\textit{Questions}} \\
  \quad \# (pre)                 & 9,428   & 9,000   \\
  \quad \# (post)                & 9,411   & 8,960   \\
  \quad Removed $\Delta$ (\%)    & 17 (0.18\%) & 40 (0.44\%) \\
  \addlinespace[1pt]
  \multicolumn{3}{@{}l}{\textit{Labeled sets}} \\
  \quad Imbalanced size          & 82,640  & 50,073  \\
  \quad \,\,\,\,Class split (\% inc / cor)
                                 & 41.37 / 58.63
                                 & 30.60 / 69.40 \\
  \quad Balanced size (50/50)    & 30,686  & 17,834  \\
  \bottomrule
\end{tabular}%
}
\end{table}

\paragraph{Dataset Imbalance}
The resulting datasets exhibit class imbalance: Spider contains 69.4\% correct candidates, while BIRD is more balanced (58.63\% correct). To study its impact, we construct balanced datasets by downsampling to a 50/50 split per question. Statistics are summarized in Table~\ref{tab:dataset_stats}. Using these datasets, we address three research questions:  
(i) cross-family ORM trainability,  
(ii) the effect of dataset balancing, and  
(iii) ORM effectiveness relative to test-time baselines.

\begin{table}[t]
\centering
\caption{ORM performance on \textit{BIRD} and \textit{Spider} with the unbalanced dataset, reporting \textit{Execution Accuracy} (\%) on dev and test sets.}
\label{tab:orm_unbalanced}



\resizebox{0.95\columnwidth}{!}{%
    \begin{tabular}{lccc}
    \toprule
    \textbf{ORM Model} & \textbf{Bird dev} & \textbf{Spider dev} & \textbf{Spider test} \\
    \midrule
    Granite-3.3-2B-Instruct & 66.88 & 82.79 & 85.14 \\
    Granite-3.3-8B-Instruct & 68.25 & 84.42 & 86.91 \\
    \cmidrule(lr){1-4}
    
    Llama-3.2-1B-Instruct   & 67.67 & 84.04 & 86.31 \\
    Llama-3.1-8B-Instruct   & 68.32 & 82.30 & 85.89 \\
    \cmidrule(lr){1-4}
    
    Qwen2.5-1.5B-Instruct   & 68.12 & 84.33 & 86.17 \\
    Qwen2.5-7B-Instruct     & 68.58 & 84.33 & 86.91 \\
    \cmidrule(lr){1-4}
    
    OmniSQL-7B              & 68.64 & 84.42 & 86.63 \\
    \bottomrule
    \end{tabular}
}
\end{table}


\noindent\textbf{Cross-Family Trainability}
Table~\ref{tab:orm_unbalanced} reports ORM performance trained on unbalanced data across all backbones. Results are tightly clustered: BIRD dev ranges from 66.88–68.64, Spider dev from 82.30–84.42, and Spider test from 85.14–86.91. Performance differences across model families and scales are modest, indicating that ORMs train reliably across heterogeneous architectures. Larger models yield only minor gains, suggesting ORM effectiveness is not strongly dependent on parameter count. Rankings are stable across datasets and splits, demonstrating robustness and family-agnostic generalization.

\begin{table}[t]
\centering
\caption{Execution accuracy (\%) of ORM models on \textit{BIRD dev}, \textit{Spider dev}, and \textit{Spider test}. Each cell shows results on \emph{Unbalanced} and \emph{Balanced} sets with $\Delta$ (green $\uparrow$ = gain, red $\downarrow$ = drop). Ordered by BIRD Unbalanced; best in \textbf{bold}, runner-up \underline{underlined}.}
\label{tab:orm_comparison_v2_v1_compact}

\resizebox{0.48\textwidth}{!}{%
\begin{tabular}{lccc}
\toprule
\textbf{ORM Model} &
\makecell{\textbf{BIRD dev}\\{\small Unbalanced $\rightarrow$ Balanced}} &
\makecell{\textbf{Spider dev}\\{\small Unbalanced $\rightarrow$ Balanced}} &
\makecell{\textbf{Spider test}\\{\small Unbalanced $\rightarrow$ Balanced}} \\
\midrule

Granite-3.3-2B-Instruct &
\makecell{66.88 $\rightarrow$ 66.17 \\ {\small\textcolor{BrickRed}{(-0.71 $\downarrow$)}}} &
\makecell{82.79 $\rightarrow$ 82.79 \\ {\small(0.00)}} &
\makecell{85.14 $\rightarrow$ 85.14 \\ {\small(0.00)}} \\

\cmidrule(lr){2-4}

Llama-3.2-1B-Instruct &
\makecell{67.67 $\rightarrow$ 67.21 \\ {\small\textcolor{BrickRed}{(-0.46 $\downarrow$)}}} &
\makecell{84.04 $\rightarrow$ 83.56 \\ {\small\textcolor{BrickRed}{(-0.48 $\downarrow$)}}} &
\makecell{86.31 $\rightarrow$ 86.54 \\ {\small\textcolor{ForestGreen}{(+0.23 $\uparrow$)}}} \\

\cmidrule(lr){2-4}

Qwen2.5-1.5B-Instruct &
\makecell{68.12 $\rightarrow$ 67.28 \\ {\small\textcolor{BrickRed}{(-0.84 $\downarrow$)}}} &
\makecell{84.33 $\rightarrow$ 83.56 \\ {\small\textcolor{BrickRed}{(-0.77 $\downarrow$)}}} &
\makecell{86.17 $\rightarrow$ 86.31 \\ {\small\textcolor{ForestGreen}{(+0.14 $\uparrow$)}}} \\

\cmidrule(lr){2-4}

Granite-3.3-8B-Instruct &
\makecell{68.25 $\rightarrow$ 68.38 \\ {\small\textcolor{ForestGreen}{(+0.13 $\uparrow$)}}} &
\makecell{84.42 $\rightarrow$ 83.85 \\ {\small\textcolor{BrickRed}{(-0.57 $\downarrow$)}}} &
\makecell{86.91 $\rightarrow$ \underline{87.00} \\ {\small\textcolor{ForestGreen}{(+0.09 $\uparrow$)}}} \\

\cmidrule(lr){2-4}

Llama-3.1-8B-Instruct &
\makecell{68.32 $\rightarrow$ 67.86 \\ {\small\textcolor{BrickRed}{(-0.46 $\downarrow$)}}} &
\makecell{82.30 $\rightarrow$ 84.11 \\ {\small\textcolor{ForestGreen}{(+1.81 $\uparrow$)}}} &
\makecell{85.89 $\rightarrow$ 86.96 \\ {\small\textcolor{ForestGreen}{(+1.07 $\uparrow$)}}} \\

\cmidrule(lr){2-4}

Qwen2.5-7B-Instruct &
\makecell{68.58 $\rightarrow$ 68.19 \\ {\small\textcolor{BrickRed}{(-0.39 $\downarrow$)}}} &
\makecell{84.33 $\rightarrow$ 84.14 \\ {\small\textcolor{BrickRed}{(-0.19 $\downarrow$)}}} &
\makecell{86.91 $\rightarrow$ 86.68 \\ {\small\textcolor{BrickRed}{(-0.23 $\downarrow$)}}} \\

\cmidrule(lr){2-4}

OmniSQL-7B &
\makecell{\underline{68.64} $\rightarrow$ \textbf{68.90} \\ {\small\textcolor{ForestGreen}{(+0.26 $\uparrow$)}}} &
\makecell{\underline{84.42} $\rightarrow$ \textbf{84.53} \\ {\small\textcolor{ForestGreen}{(+0.11 $\uparrow$)}}} &
\makecell{86.63 $\rightarrow$ \textbf{87.47} \\ {\small\textcolor{ForestGreen}{(+0.84 $\uparrow$)}}} \\

\bottomrule
\end{tabular}
}
\vspace{-6pt}
\end{table}

\noindent\textbf{Effect of Dataset Balancing}
Table~\ref{tab:orm_comparison_v2_v1_compact} compares ORMs trained on unbalanced versus balanced data. Balancing produces mixed effects on dev sets but yields more consistent improvements on Spider test, particularly for Llama-3.1-8B (+1.07) and OmniSQL-7B (+0.84). Overall changes remain small (within $\pm1.2$), indicating ORM stability under skewed label distributions.

Importantly, balanced datasets substantially reduce training size (from 80k$\rightarrow$30k for BIRD and 50k$\rightarrow$18k for Spider), lowering training cost without degrading performance. In several cases, balanced training even improves accuracy. We therefore adopt the balanced configuration for subsequent comparisons.

\paragraph{Summary}
ORMs (i) train reliably across model families, (ii) remain stable under label imbalance while benefiting from balanced data, and (iii) consistently outperform widely used test-time heuristics. These results establish ORM-based selection as a stronger and more principled alternative to Majority Voting and execution-based Best-of-$N$ for test-time inference in Text-to-SQL.

\section{Ablation Studies}\label{sec:ablations}
We conduct ablations to better understand the behavior of Outcome Reward Models (ORMs) under different design and scaling choices. In particular, we study:
(i) the effect of candidate pool size $N$,
(ii) ORM performance with larger generators,
(iii) sensitivity to training prompts,
(iv) scaling ORMs beyond 7B parameters, and
(v) the impact of fine-tuning objectives.

\begin{table*}[t]
\centering
\caption{Execution accuracy (\%) and Pass@$N$ of test-time strategies (execution-based BoN, Majority Voting, ORM-based BoN) across varying $N$ values on the \textit{BIRD dev}, \textit{Spider dev}, and \textit{Spider test} benchmarks. Columns are color-coded from light yellow (lowest) to dark orange (highest) performance.}
\label{tab:n_decreasing}

\renewcommand{\arraystretch}{1.15} 

\resizebox{0.95\textwidth}{!}{%
\begin{tabular}{c
                ccc
                ccc
                ccc
                ccc}
    \toprule
    \multirow{2}{*}{$N$} 
    & \multicolumn{3}{c}{\textbf{Execution-based BoN}}
    & \multicolumn{3}{c}{\textbf{Majority Voting}} 
    & \multicolumn{3}{c}{\textbf{ORM-based BoN}} 
    & \multicolumn{3}{c}{\textbf{Pass@$N$}} \\ 
    \cmidrule(lr){2-4}\cmidrule(lr){5-7}\cmidrule(lr){8-10}\cmidrule(lr){11-13}
    & BIRD dev & Spider dev & Spider test 
    & BIRD dev & Spider dev & Spider test 
    & BIRD dev & Spider dev & Spider test 
    & BIRD dev & Spider dev & Spider test \\
    \midrule
    32 & \cellcolor{heatmax!100!heatmin} 66.04 & \cellcolor{heatmax!81!heatmin} 82.79 & \cellcolor{heatmax!100!heatmin} 85.14 & \cellcolor{heatmax!94!heatmin} 66.95 & \cellcolor{heatmax!78!heatmin} 83.75 & \cellcolor{heatmax!84!heatmin} 85.47 & \cellcolor{heatmax!100!heatmin} 68.90 & \cellcolor{heatmax!92!heatmin} 84.53 & \cellcolor{heatmax!100!heatmin} 87.47 & \cellcolor{heatmax!100!heatmin} 80.57 & \cellcolor{heatmax!100!heatmin} 91.68 & \cellcolor{heatmax!100!heatmin} 93.29 \\
    31 & \cellcolor{heatmax!100!heatmin} 66.04 & \cellcolor{heatmax!81!heatmin} 82.79 & \cellcolor{heatmax!100!heatmin} 85.14 & \cellcolor{heatmax!98!heatmin} 67.08 & \cellcolor{heatmax!78!heatmin} 83.75 & \cellcolor{heatmax!84!heatmin} 85.47 & \cellcolor{heatmax!94!heatmin} 68.58 & \cellcolor{heatmax!92!heatmin} 84.53 & \cellcolor{heatmax!100!heatmin} 87.47 & \cellcolor{heatmax!98!heatmin} 80.25 & \cellcolor{heatmax!100!heatmin} 91.68 & \cellcolor{heatmax!99!heatmin} 93.20 \\
    30 & \cellcolor{heatmax!100!heatmin} 66.04 & \cellcolor{heatmax!100!heatmin} 82.88 & \cellcolor{heatmax!100!heatmin} 85.14 & \cellcolor{heatmax!96!heatmin} 67.01 & \cellcolor{heatmax!72!heatmin} 83.66 & \cellcolor{heatmax!81!heatmin} 85.42 & \cellcolor{heatmax!94!heatmin} 68.58 & \cellcolor{heatmax!96!heatmin} 84.62 & \cellcolor{heatmax!100!heatmin} 87.47 & \cellcolor{heatmax!98!heatmin} 80.25 & \cellcolor{heatmax!100!heatmin} 91.68 & \cellcolor{heatmax!98!heatmin} 93.15 \\
    29 & \cellcolor{heatmax!97!heatmin} 65.97 & \cellcolor{heatmax!100!heatmin} 82.88 & \cellcolor{heatmax!100!heatmin} 85.14 & \cellcolor{heatmax!94!heatmin} 66.95 & \cellcolor{heatmax!67!heatmin} 83.56 & \cellcolor{heatmax!81!heatmin} 85.42 & \cellcolor{heatmax!92!heatmin} 68.51 & \cellcolor{heatmax!96!heatmin} 84.62 & \cellcolor{heatmax!99!heatmin} 87.42 & \cellcolor{heatmax!97!heatmin} 80.12 & \cellcolor{heatmax!100!heatmin} 91.68 & \cellcolor{heatmax!98!heatmin} 93.11 \\
    28 & \cellcolor{heatmax!97!heatmin} 65.97 & \cellcolor{heatmax!100!heatmin} 82.88 & \cellcolor{heatmax!100!heatmin} 85.14 & \cellcolor{heatmax!96!heatmin} 67.01 & \cellcolor{heatmax!78!heatmin} 83.75 & \cellcolor{heatmax!84!heatmin} 85.47 & \cellcolor{heatmax!94!heatmin} 68.58 & \cellcolor{heatmax!96!heatmin} 84.62 & \cellcolor{heatmax!97!heatmin} 87.38 & \cellcolor{heatmax!97!heatmin} 80.05 & \cellcolor{heatmax!100!heatmin} 91.68 & \cellcolor{heatmax!97!heatmin} 93.01 \\
    27 & \cellcolor{heatmax!97!heatmin} 65.97 & \cellcolor{heatmax!100!heatmin} 82.88 & \cellcolor{heatmax!100!heatmin} 85.14 & \cellcolor{heatmax!100!heatmin} 67.14 & \cellcolor{heatmax!83!heatmin} 83.85 & \cellcolor{heatmax!84!heatmin} 85.47 & \cellcolor{heatmax!94!heatmin} 68.58 & \cellcolor{heatmax!96!heatmin} 84.62 & \cellcolor{heatmax!97!heatmin} 87.38 & \cellcolor{heatmax!96!heatmin} 79.92 & \cellcolor{heatmax!99!heatmin} 91.59 & \cellcolor{heatmax!97!heatmin} 93.01 \\
    26 & \cellcolor{heatmax!97!heatmin} 65.97 & \cellcolor{heatmax!100!heatmin} 82.88 & \cellcolor{heatmax!100!heatmin} 85.14 & \cellcolor{heatmax!88!heatmin} 66.75 & \cellcolor{heatmax!78!heatmin} 83.75 & \cellcolor{heatmax!76!heatmin} 85.33 & \cellcolor{heatmax!95!heatmin} 68.64 & \cellcolor{heatmax!96!heatmin} 84.62 & \cellcolor{heatmax!97!heatmin} 87.38 & \cellcolor{heatmax!96!heatmin} 79.92 & \cellcolor{heatmax!99!heatmin} 91.59 & \cellcolor{heatmax!95!heatmin} 92.87 \\
    25 & \cellcolor{heatmax!97!heatmin} 65.97 & \cellcolor{heatmax!100!heatmin} 82.88 & \cellcolor{heatmax!100!heatmin} 85.14 & \cellcolor{heatmax!88!heatmin} 66.75 & \cellcolor{heatmax!83!heatmin} 83.85 & \cellcolor{heatmax!86!heatmin} 85.51 & \cellcolor{heatmax!96!heatmin} 68.71 & \cellcolor{heatmax!96!heatmin} 84.62 & \cellcolor{heatmax!97!heatmin} 87.38 & \cellcolor{heatmax!95!heatmin} 79.79 & \cellcolor{heatmax!99!heatmin} 91.59 & \cellcolor{heatmax!95!heatmin} 92.87 \\
    24 & \cellcolor{heatmax!97!heatmin} 65.97 & \cellcolor{heatmax!100!heatmin} 82.88 & \cellcolor{heatmax!100!heatmin} 85.14 & \cellcolor{heatmax!96!heatmin} 67.01 & \cellcolor{heatmax!89!heatmin} 83.95 & \cellcolor{heatmax!89!heatmin} 85.56 & \cellcolor{heatmax!97!heatmin} 68.77 & \cellcolor{heatmax!96!heatmin} 84.62 & \cellcolor{heatmax!96!heatmin} 87.33 & \cellcolor{heatmax!94!heatmin} 79.53 & \cellcolor{heatmax!99!heatmin} 91.59 & \cellcolor{heatmax!94!heatmin} 92.78 \\
    23 & \cellcolor{heatmax!97!heatmin} 65.97 & \cellcolor{heatmax!100!heatmin} 82.88 & \cellcolor{heatmax!100!heatmin} 85.14 & \cellcolor{heatmax!98!heatmin} 67.08 & \cellcolor{heatmax!100!heatmin} 84.14 & \cellcolor{heatmax!84!heatmin} 85.47 & \cellcolor{heatmax!96!heatmin} 68.71 & \cellcolor{heatmax!96!heatmin} 84.62 & \cellcolor{heatmax!96!heatmin} 87.33 & \cellcolor{heatmax!93!heatmin} 79.47 & \cellcolor{heatmax!98!heatmin} 91.49 & \cellcolor{heatmax!94!heatmin} 92.78 \\
    22 & \cellcolor{heatmax!97!heatmin} 65.97 & \cellcolor{heatmax!100!heatmin} 82.88 & \cellcolor{heatmax!100!heatmin} 85.14 & \cellcolor{heatmax!92!heatmin} 66.88 & \cellcolor{heatmax!89!heatmin} 83.95 & \cellcolor{heatmax!81!heatmin} 85.42 & \cellcolor{heatmax!100!heatmin} 68.90 & \cellcolor{heatmax!96!heatmin} 84.62 & \cellcolor{heatmax!96!heatmin} 87.33 & \cellcolor{heatmax!93!heatmin} 79.33 & \cellcolor{heatmax!97!heatmin} 91.39 & \cellcolor{heatmax!94!heatmin} 92.73 \\
    21 & \cellcolor{heatmax!94!heatmin} 65.91 & \cellcolor{heatmax!100!heatmin} 82.88 & \cellcolor{heatmax!100!heatmin} 85.14 & \cellcolor{heatmax!88!heatmin} 66.75 & \cellcolor{heatmax!83!heatmin} 83.85 & \cellcolor{heatmax!84!heatmin} 85.47 & \cellcolor{heatmax!99!heatmin} 68.84 & \cellcolor{heatmax!96!heatmin} 84.62 & \cellcolor{heatmax!94!heatmin} 87.28 & \cellcolor{heatmax!91!heatmin} 79.14 & \cellcolor{heatmax!95!heatmin} 91.20 & \cellcolor{heatmax!94!heatmin} 92.73 \\
    20 & \cellcolor{heatmax!94!heatmin} 65.91 & \cellcolor{heatmax!100!heatmin} 82.88 & \cellcolor{heatmax!100!heatmin} 85.14 & \cellcolor{heatmax!96!heatmin} 67.01 & \cellcolor{heatmax!78!heatmin} 83.75 & \cellcolor{heatmax!89!heatmin} 85.56 & \cellcolor{heatmax!97!heatmin} 68.77 & \cellcolor{heatmax!96!heatmin} 84.62 & \cellcolor{heatmax!96!heatmin} 87.33 & \cellcolor{heatmax!91!heatmin} 79.07 & \cellcolor{heatmax!93!heatmin} 91.01 & \cellcolor{heatmax!94!heatmin} 92.69 \\
    19 & \cellcolor{heatmax!94!heatmin} 65.91 & \cellcolor{heatmax!100!heatmin} 82.88 & \cellcolor{heatmax!100!heatmin} 85.14 & \cellcolor{heatmax!98!heatmin} 67.08 & \cellcolor{heatmax!83!heatmin} 83.85 & \cellcolor{heatmax!78!heatmin} 85.37 & \cellcolor{heatmax!97!heatmin} 68.77 & \cellcolor{heatmax!96!heatmin} 84.62 & \cellcolor{heatmax!96!heatmin} 87.33 & \cellcolor{heatmax!90!heatmin} 78.88 & \cellcolor{heatmax!92!heatmin} 90.91 & \cellcolor{heatmax!92!heatmin} 92.59 \\
    18 & \cellcolor{heatmax!94!heatmin} 65.91 & \cellcolor{heatmax!100!heatmin} 82.88 & \cellcolor{heatmax!100!heatmin} 85.14 & \cellcolor{heatmax!96!heatmin} 67.01 & \cellcolor{heatmax!78!heatmin} 83.76 & \cellcolor{heatmax!94!heatmin} 85.65 & \cellcolor{heatmax!99!heatmin} 68.84 & \cellcolor{heatmax!96!heatmin} 84.62 & \cellcolor{heatmax!97!heatmin} 87.38 & \cellcolor{heatmax!90!heatmin} 78.88 & \cellcolor{heatmax!91!heatmin} 90.81 & \cellcolor{heatmax!92!heatmin} 92.55 \\
    17 & \cellcolor{heatmax!94!heatmin} 65.91 & \cellcolor{heatmax!100!heatmin} 82.88 & \cellcolor{heatmax!100!heatmin} 85.14 & \cellcolor{heatmax!98!heatmin} 67.08 & \cellcolor{heatmax!89!heatmin} 83.95 & \cellcolor{heatmax!92!heatmin} 85.61 & \cellcolor{heatmax!97!heatmin} 68.77 & \cellcolor{heatmax!100!heatmin} 84.72 & \cellcolor{heatmax!96!heatmin} 87.33 & \cellcolor{heatmax!89!heatmin} 78.68 & \cellcolor{heatmax!91!heatmin} 90.81 & \cellcolor{heatmax!91!heatmin} 92.50 \\
    16 & \cellcolor{heatmax!94!heatmin} 65.91 & \cellcolor{heatmax!100!heatmin} 82.88 & \cellcolor{heatmax!100!heatmin} 85.14 & \cellcolor{heatmax!94!heatmin} 66.95 & \cellcolor{heatmax!72!heatmin} 83.66 & \cellcolor{heatmax!92!heatmin} 85.61 & \cellcolor{heatmax!97!heatmin} 68.77 & \cellcolor{heatmax!92!heatmin} 84.53 & \cellcolor{heatmax!96!heatmin} 87.33 & \cellcolor{heatmax!88!heatmin} 78.49 & \cellcolor{heatmax!90!heatmin} 90.72 & \cellcolor{heatmax!90!heatmin} 92.40 \\
    15 & \cellcolor{heatmax!91!heatmin} 65.84 & \cellcolor{heatmax!100!heatmin} 82.88 & \cellcolor{heatmax!100!heatmin} 85.14 & \cellcolor{heatmax!100!heatmin} 67.14 & \cellcolor{heatmax!83!heatmin} 83.85 & \cellcolor{heatmax!81!heatmin} 85.42 & \cellcolor{heatmax!96!heatmin} 68.71 & \cellcolor{heatmax!92!heatmin} 84.53 & \cellcolor{heatmax!94!heatmin} 87.28 & \cellcolor{heatmax!86!heatmin} 78.23 & \cellcolor{heatmax!87!heatmin} 90.52 & \cellcolor{heatmax!89!heatmin} 92.27 \\
    14 & \cellcolor{heatmax!91!heatmin} 65.84 & \cellcolor{heatmax!100!heatmin} 82.88 & \cellcolor{heatmax!96!heatmin} 85.10 & \cellcolor{heatmax!96!heatmin} 67.01 & \cellcolor{heatmax!83!heatmin} 83.85 & \cellcolor{heatmax!92!heatmin} 85.61 & \cellcolor{heatmax!95!heatmin} 68.64 & \cellcolor{heatmax!92!heatmin} 84.53 & \cellcolor{heatmax!92!heatmin} 87.19 & \cellcolor{heatmax!84!heatmin} 77.90 & \cellcolor{heatmax!87!heatmin} 90.52 & \cellcolor{heatmax!87!heatmin} 92.13 \\
    13 & \cellcolor{heatmax!91!heatmin} 65.84 & \cellcolor{heatmax!100!heatmin} 82.88 & \cellcolor{heatmax!96!heatmin} 85.10 & \cellcolor{heatmax!100!heatmin} 67.14 & \cellcolor{heatmax!72!heatmin} 83.66 & \cellcolor{heatmax!89!heatmin} 85.56 & \cellcolor{heatmax!95!heatmin} 68.64 & \cellcolor{heatmax!92!heatmin} 84.53 & \cellcolor{heatmax!92!heatmin} 87.19 & \cellcolor{heatmax!84!heatmin} 77.84 & \cellcolor{heatmax!87!heatmin} 90.43 & \cellcolor{heatmax!87!heatmin} 92.08 \\
    12 & \cellcolor{heatmax!91!heatmin} 65.84 & \cellcolor{heatmax!100!heatmin} 82.88 & \cellcolor{heatmax!92!heatmin} 85.05 & \cellcolor{heatmax!92!heatmin} 66.88 & \cellcolor{heatmax!83!heatmin} 83.85 & \cellcolor{heatmax!86!heatmin} 85.51 & \cellcolor{heatmax!96!heatmin} 68.71 & \cellcolor{heatmax!88!heatmin} 84.44 & \cellcolor{heatmax!85!heatmin} 86.96 & \cellcolor{heatmax!83!heatmin} 77.71 & \cellcolor{heatmax!87!heatmin} 90.43 & \cellcolor{heatmax!85!heatmin} 91.94 \\
    11 & \cellcolor{heatmax!91!heatmin} 65.84 & \cellcolor{heatmax!100!heatmin} 82.88 & \cellcolor{heatmax!92!heatmin} 85.05 & \cellcolor{heatmax!90!heatmin} 66.82 & \cellcolor{heatmax!72!heatmin} 83.66 & \cellcolor{heatmax!68!heatmin} 85.19 & \cellcolor{heatmax!95!heatmin} 68.64 & \cellcolor{heatmax!83!heatmin} 84.33 & \cellcolor{heatmax!83!heatmin} 86.87 & \cellcolor{heatmax!80!heatmin} 77.25 & \cellcolor{heatmax!84!heatmin} 90.23 & \cellcolor{heatmax!83!heatmin} 91.71 \\
    10 & \cellcolor{heatmax!85!heatmin} 65.71 & \cellcolor{heatmax!100!heatmin} 82.88 & \cellcolor{heatmax!92!heatmin} 85.05 & \cellcolor{heatmax!86!heatmin} 66.69 & \cellcolor{heatmax!100!heatmin} 84.14 & \cellcolor{heatmax!89!heatmin} 85.56 & \cellcolor{heatmax!95!heatmin} 68.64 & \cellcolor{heatmax!83!heatmin} 84.33 & \cellcolor{heatmax!84!heatmin} 86.91 & \cellcolor{heatmax!79!heatmin} 76.99 & \cellcolor{heatmax!82!heatmin} 90.04 & \cellcolor{heatmax!82!heatmin} 91.66 \\
    9 & \cellcolor{heatmax!85!heatmin} 65.71 & \cellcolor{heatmax!100!heatmin} 82.88 & \cellcolor{heatmax!92!heatmin} 85.05 & \cellcolor{heatmax!80!heatmin} 66.49 & \cellcolor{heatmax!94!heatmin} 84.04 & \cellcolor{heatmax!94!heatmin} 85.65 & \cellcolor{heatmax!92!heatmin} 68.51 & \cellcolor{heatmax!79!heatmin} 84.24 & \cellcolor{heatmax!84!heatmin} 86.91 & \cellcolor{heatmax!76!heatmin} 76.60 & \cellcolor{heatmax!80!heatmin} 89.85 & \cellcolor{heatmax!82!heatmin} 91.62 \\
    8 & \cellcolor{heatmax!85!heatmin} 65.71 & \cellcolor{heatmax!100!heatmin} 82.88 & \cellcolor{heatmax!92!heatmin} 85.05 & \cellcolor{heatmax!94!heatmin} 66.95 & \cellcolor{heatmax!89!heatmin} 83.95 & \cellcolor{heatmax!92!heatmin} 85.61 & \cellcolor{heatmax!92!heatmin} 68.51 & \cellcolor{heatmax!83!heatmin} 84.33 & \cellcolor{heatmax!81!heatmin} 86.82 & \cellcolor{heatmax!73!heatmin} 76.14 & \cellcolor{heatmax!78!heatmin} 89.65 & \cellcolor{heatmax!80!heatmin} 91.48 \\
    7 & \cellcolor{heatmax!79!heatmin} 65.58 & \cellcolor{heatmax!81!heatmin} 82.79 & \cellcolor{heatmax!92!heatmin} 85.05 & \cellcolor{heatmax!84!heatmin} 66.62 & \cellcolor{heatmax!94!heatmin} 84.04 & \cellcolor{heatmax!97!heatmin} 85.70 & \cellcolor{heatmax!90!heatmin} 68.38 & \cellcolor{heatmax!83!heatmin} 84.33 & \cellcolor{heatmax!80!heatmin} 86.77 & \cellcolor{heatmax!69!heatmin} 75.36 & \cellcolor{heatmax!74!heatmin} 89.26 & \cellcolor{heatmax!76!heatmin} 91.06 \\
    6 & \cellcolor{heatmax!73!heatmin} 65.45 & \cellcolor{heatmax!81!heatmin} 82.79 & \cellcolor{heatmax!88!heatmin} 85.00 & \cellcolor{heatmax!78!heatmin} 66.43 & \cellcolor{heatmax!72!heatmin} 83.66 & \cellcolor{heatmax!100!heatmin} 85.75 & \cellcolor{heatmax!79!heatmin} 67.86 & \cellcolor{heatmax!79!heatmin} 84.24 & \cellcolor{heatmax!73!heatmin} 86.54 & \cellcolor{heatmax!64!heatmin} 74.58 & \cellcolor{heatmax!73!heatmin} 89.17 & \cellcolor{heatmax!73!heatmin} 90.82 \\
    5 & \cellcolor{heatmax!69!heatmin} 65.38 & \cellcolor{heatmax!81!heatmin} 82.79 & \cellcolor{heatmax!88!heatmin} 85.00 & \cellcolor{heatmax!64!heatmin} 65.97 & \cellcolor{heatmax!89!heatmin} 83.95 & \cellcolor{heatmax!94!heatmin} 85.65 & \cellcolor{heatmax!71!heatmin} 67.47 & \cellcolor{heatmax!75!heatmin} 84.14 & \cellcolor{heatmax!72!heatmin} 86.49 & \cellcolor{heatmax!59!heatmin} 73.79 & \cellcolor{heatmax!67!heatmin} 88.59 & \cellcolor{heatmax!70!heatmin} 90.54 \\
    4 & \cellcolor{heatmax!67!heatmin} 65.32 & \cellcolor{heatmax!100!heatmin} 82.88 & \cellcolor{heatmax!88!heatmin} 85.00 & \cellcolor{heatmax!44!heatmin} 65.32 & \cellcolor{heatmax!72!heatmin} 83.66 & \cellcolor{heatmax!84!heatmin} 85.47 & \cellcolor{heatmax!69!heatmin} 67.34 & \cellcolor{heatmax!71!heatmin} 84.04 & \cellcolor{heatmax!72!heatmin} 86.49 & \cellcolor{heatmax!54!heatmin} 72.82 & \cellcolor{heatmax!64!heatmin} 88.30 & \cellcolor{heatmax!66!heatmin} 90.13 \\
    3 & \cellcolor{heatmax!57!heatmin} 65.12 & \cellcolor{heatmax!100!heatmin} 82.88 & \cellcolor{heatmax!79!heatmin} 84.91 & \cellcolor{heatmax!40!heatmin} 65.19 & \cellcolor{heatmax!67!heatmin} 83.56 & \cellcolor{heatmax!76!heatmin} 85.33 & \cellcolor{heatmax!57!heatmin} 66.75 & \cellcolor{heatmax!71!heatmin} 84.04 & \cellcolor{heatmax!62!heatmin} 86.17 & \cellcolor{heatmax!42!heatmin} 70.86 & \cellcolor{heatmax!54!heatmin} 87.43 & \cellcolor{heatmax!55!heatmin} 89.10 \\
    2 & \cellcolor{heatmax!45!heatmin} 64.86 & \cellcolor{heatmax!81!heatmin} 82.79 & \cellcolor{heatmax!54!heatmin} 84.63 & \cellcolor{heatmax!16!heatmin} 64.41 & \cellcolor{heatmax!39!heatmin} 83.08 & \cellcolor{heatmax!24!heatmin} 84.44 & \cellcolor{heatmax!43!heatmin} 66.04 & \cellcolor{heatmax!54!heatmin} 83.66 & \cellcolor{heatmax!50!heatmin} 85.75 & \cellcolor{heatmax!30!heatmin} 68.84 & \cellcolor{heatmax!40!heatmin} 86.07 & \cellcolor{heatmax!39!heatmin} 87.66 \\
    1 & \cellcolor{heatmax!0!heatmin} 63.89 & \cellcolor{heatmax!0!heatmin} 82.40 & \cellcolor{heatmax!0!heatmin} 84.02 & \cellcolor{heatmax!0!heatmin} 63.89 & \cellcolor{heatmax!0!heatmin} 82.40 & \cellcolor{heatmax!0!heatmin} 84.02 & \cellcolor{heatmax!0!heatmin} 63.89 & \cellcolor{heatmax!0!heatmin} 82.40 & \cellcolor{heatmax!0!heatmin} 84.02 & \cellcolor{heatmax!0!heatmin} 63.89 & \cellcolor{heatmax!0!heatmin} 82.40 & \cellcolor{heatmax!0!heatmin} 84.02 \\
    \bottomrule
    \end{tabular}
}
\end{table*}

\begin{figure}[t]
  \centering
  \resizebox{\columnwidth}{!}{%
    \begin{minipage}{\columnwidth}
      \begin{minipage}[t]{0.48\columnwidth}
        \centering
        \includegraphics[width=0.95\linewidth]{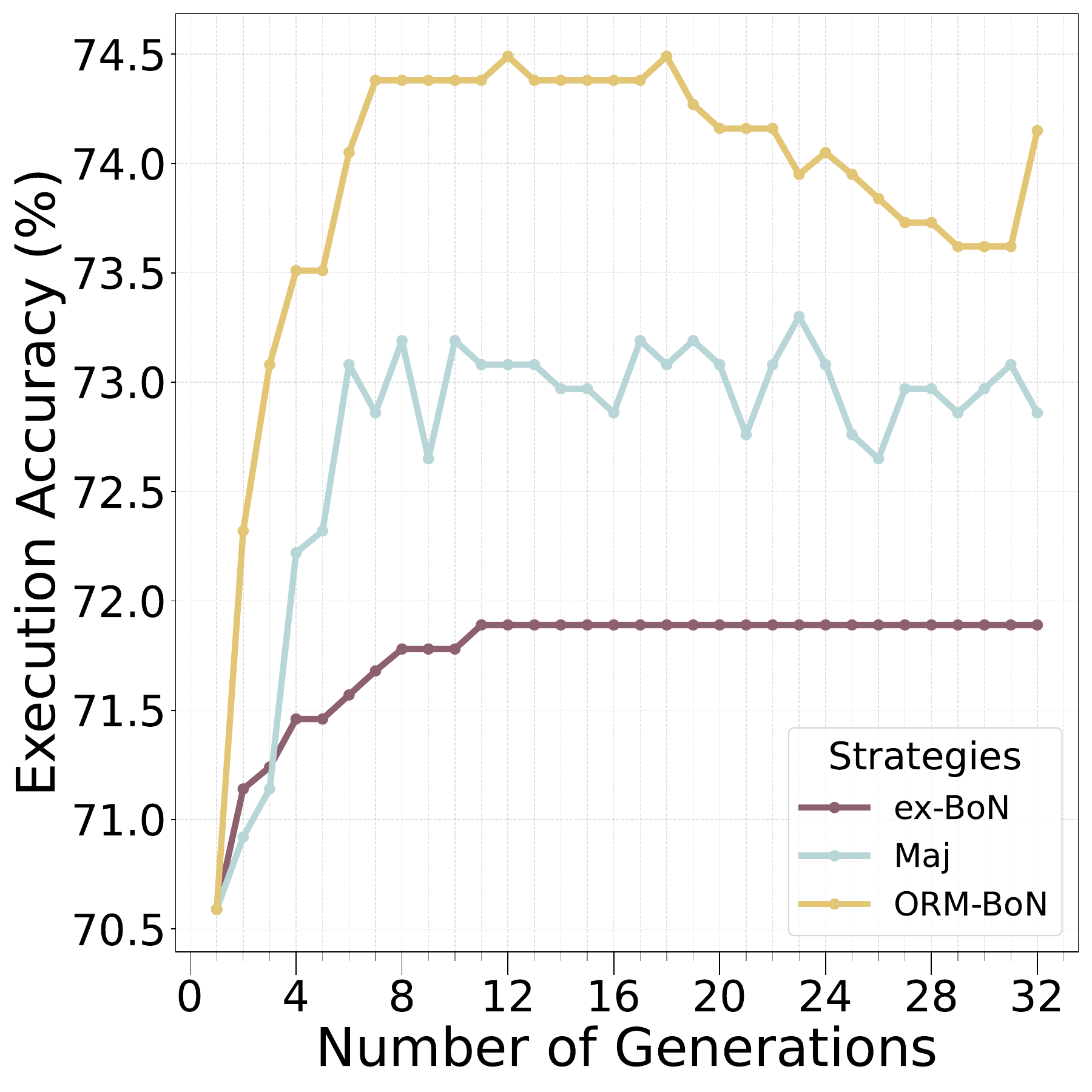}
        \par (a) Simple
      \end{minipage}\hfill
      \begin{minipage}[t]{0.48\columnwidth}
        \centering
        \includegraphics[width=0.95\linewidth]{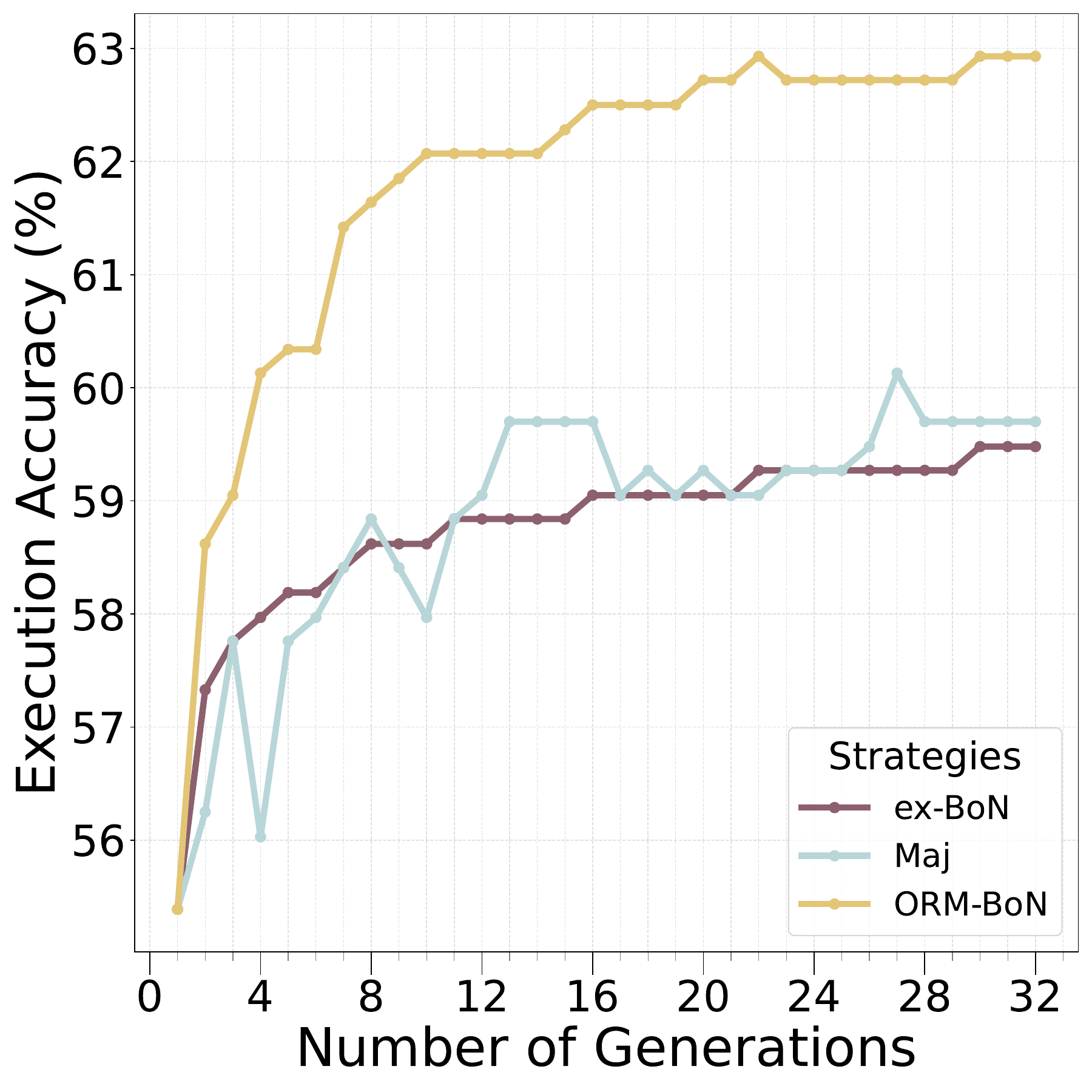}
        \par (b) Moderate
      \end{minipage}

      \vspace{0.3em}

      \begin{minipage}[t]{0.48\columnwidth}
        \centering
        \includegraphics[width=0.95\linewidth]{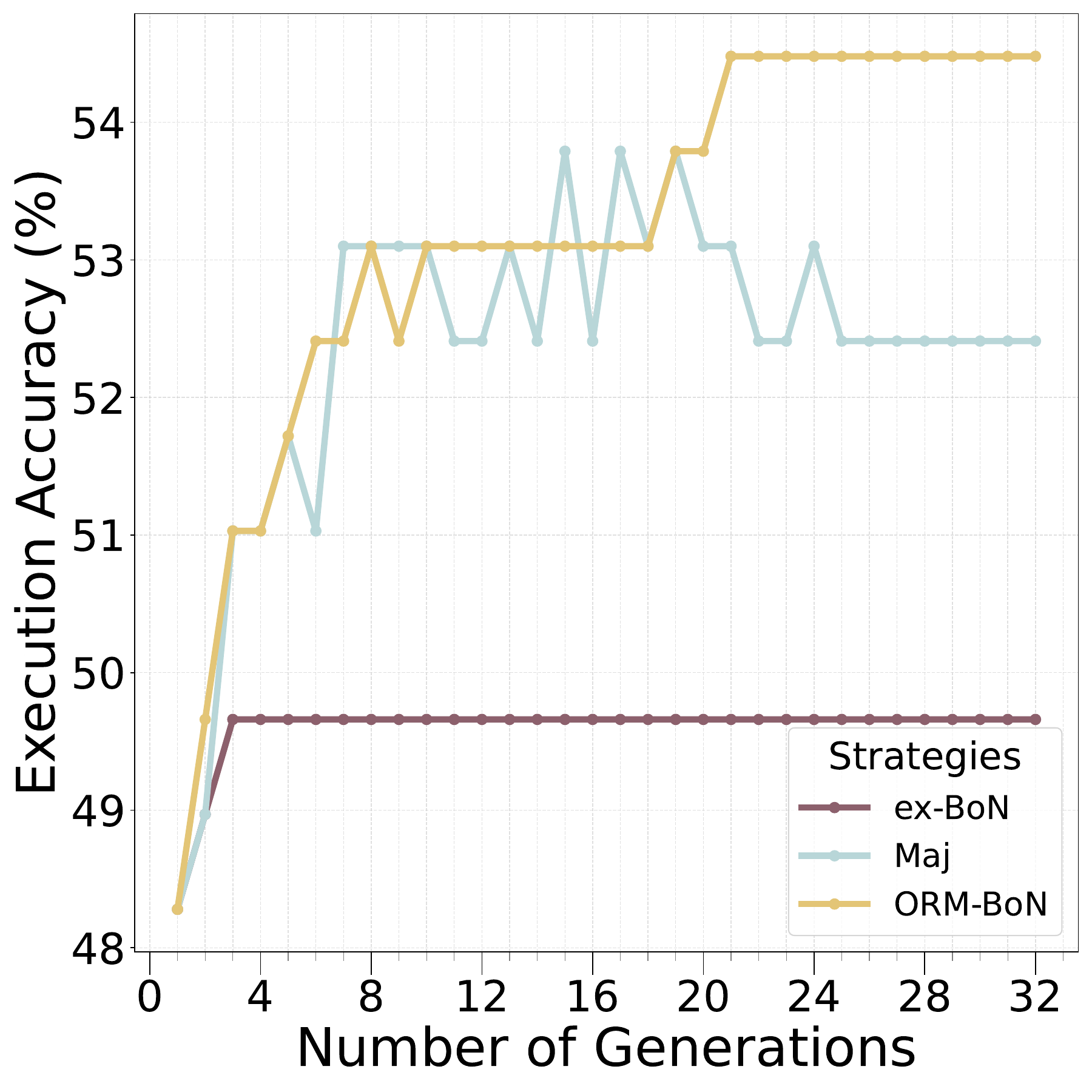}
        \par (c) Challenging
      \end{minipage}\hfill
      \begin{minipage}[t]{0.48\columnwidth}
        \centering
        \includegraphics[width=0.95\linewidth]{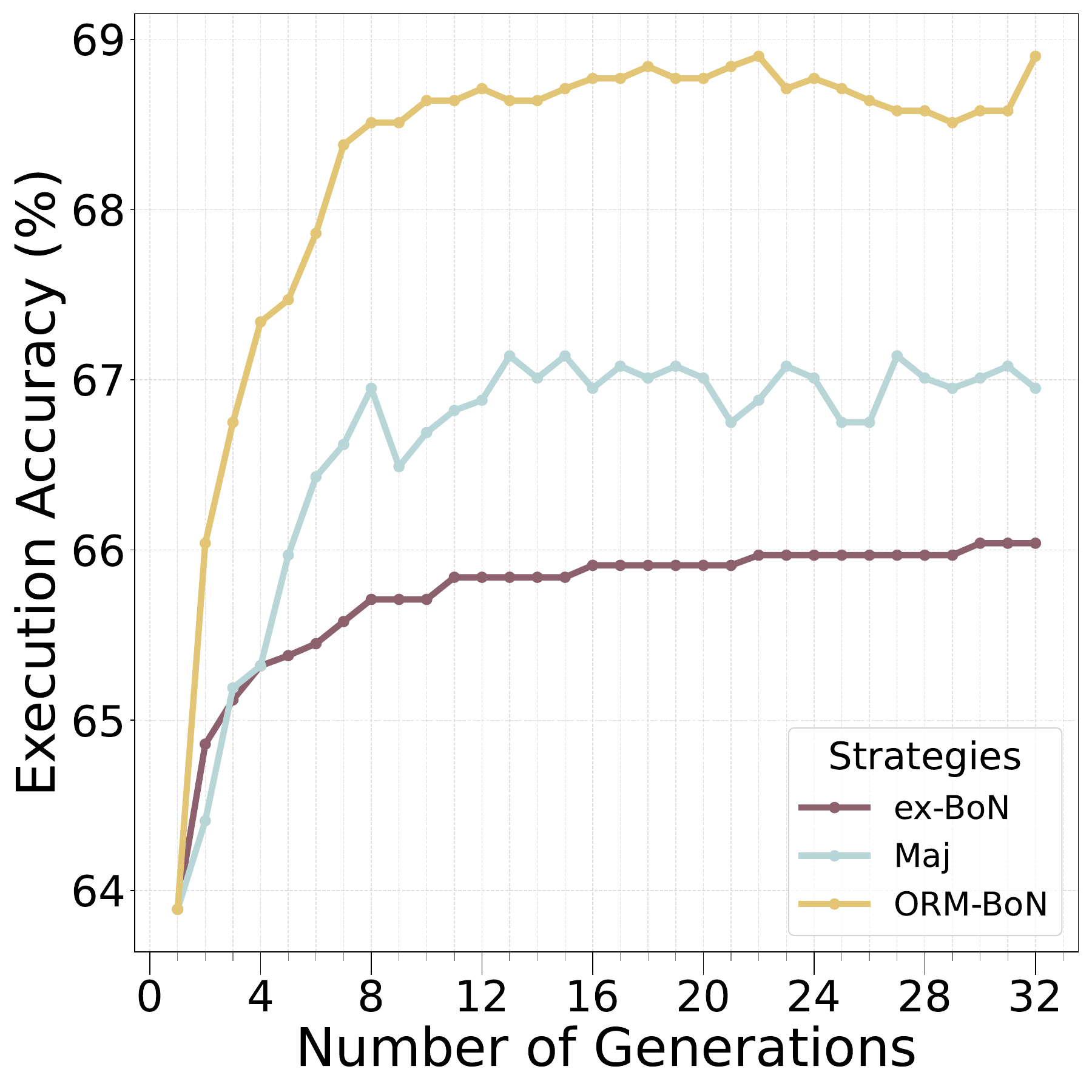}
        \par (d) Overall
      \end{minipage}
    \end{minipage}%
  }
  \caption{Execution accuracy on \textit{BIRD dev} as a function of $N$, comparing execution-based Best-of-$N$, Majority Voting, and ORM-based Best-of-$N$, stratified by query difficulty.}
  \label{fig:n_decreasing_difficulty}
\end{figure}

\paragraph{Effect of Candidate Pool Size $N$}
Table~\ref{tab:n_decreasing} shows that ORM-based selection consistently outperforms execution-based Best-of-$N$ and Majority Voting across all values of $N$ on \textit{BIRD} and \textit{Spider}. While heuristic methods saturate early, ORM performance continues to improve as $N$ increases and more closely approaches the oracle upper bound defined by Pass@$N$. Gains are largest on challenging queries (Figure~\ref{fig:n_decreasing_difficulty}), where semantic discrimination is most critical. These results indicate that ORMs more effectively leverage increased test-time compute.

\begin{table}[!t]
\centering
\caption{Execution accuracy (\%) and Pass@32 on the \textit{BIRD dev}, \textit{Spider dev}, and \textit{Spider test}. Candidate queries are generated by OmniSQL models with 7B, 14B, and 32B parameters, while the ORM is implemented using OmniSQL-7B. Best scores per column are shown in \textbf{bold}.}
\label{tab:generator_comparison}


\resizebox{0.95\columnwidth}{!}{%
  \begin{tabular}{c
                  l
                  S[table-format=2.2]
                  S[table-format=2.2]
                  S[table-format=2.2]
                  S[table-format=2.2]
                  S[table-format=2.2]}
    \toprule
    & \textbf{Model} & {Baseline} & {Ex-BoN} & {Majority Voting} & {ORM-BoN} & {Pass@32} \\
    \midrule
    \multirow{3}{*}{\rotatebox{90}{\parbox{0.8cm}{\centering \scriptsize \textbf{BIRD dev}}}} 
      & OmniSQL-7B  & 63.89 & 66.04 & 66.95 & 68.90 & 80.57 \\
      & OmniSQL-14B & \bfseries 64.28 & \bfseries 66.75 & \bfseries 67.01 & 69.04 & 80.90 \\
      & OmniSQL-32B & 64.02 & 66.10 & 63.95 & \bfseries 69.30 & \bfseries 82.20 \\
    \midrule
    \multirow{3}{*}{\rotatebox{90}{\parbox{0.8cm}{\centering \scriptsize \textbf{Spider dev}}}} 
      & OmniSQL-7B  & 82.40 & 82.79 & \bfseries 83.75 & 84.33 & 91.68 \\
      & OmniSQL-14B & 82.01 & 82.20 & 83.56 & \bfseries 84.62 & 92.75 \\
      & OmniSQL-32B & \bfseries 83.17 & \bfseries 83.37 & 83.08 & 84.43 & \bfseries 92.84 \\
    \midrule
    \multirow{3}{*}{\rotatebox{90}{\parbox{0.8cm}{\centering \scriptsize \textbf{Spider test}}}} 
      & OmniSQL-7B  & 84.02 & 85.14 & 85.47 & 87.33 & 93.29 \\
      & OmniSQL-14B & \bfseries 84.91 & \bfseries 85.28 & 84.86 & \bfseries 87.42 & 93.53 \\
      & OmniSQL-32B & 84.30 & 84.72 & \bfseries 86.07 & 87.19 & \bfseries 94.08 \\
    \bottomrule
  \end{tabular}
}
\end{table}
\paragraph{ORMs with Larger Generators}
Table~\ref{tab:generator_comparison} evaluates ORM-based selection with increasingly large OmniSQL generators (7B, 14B, 32B). Across \textit{BIRD} and \textit{Spider}, ORMs consistently outperform heuristic baselines and maintain stable improvements as generator size increases. While larger generators improve absolute performance, the relative gains from ORM-based selection persist, indicating that ORM benefits are complementary to generator scaling.





\begin{table}[t]
\centering
\caption{Ablation study on verification prompt design for ORM on \textit{BIRD dev}. Execution accuracy (\%) by query difficulty. Generator: OmniSQL-7B; ORM: Qwen2.5-7B-Instruct. Best scores in \textbf{bold}.}
\label{tab:ablation_prompt}

\footnotesize

\resizebox{0.95\columnwidth}{!}{%
    \begin{tabular}{lcccc}
        \toprule
        \textbf{Prompt Variant} & \textbf{Simple} & \textbf{Moderate} & \textbf{Challenging} & \textbf{Total} \\
        \midrule
        Instruction & 72.65 & 59.70 & 51.32 & 66.72 \\ 
        Data-Only & 72.76 & 59.48 & 51.72 & 66.75 \\ 
        SQL-Only & 72.86 & \textbf{63.36} & \textbf{53.79} & \textbf{68.19} \\
        Data + SQL & \textbf{73.08} & 62.50 & 50.34 & 67.73 \\
        \bottomrule
    \end{tabular}
}
\end{table}

\paragraph{Impact of Training Prompt}
Table~\ref{tab:ablation_prompt} studies prompt design for ORM training. Differences are negligible for simple queries but become pronounced for moderate and challenging ones. The \textit{SQL-only} prompt consistently yields the best overall accuracy, outperforming instruction-style and data-only variants. This suggests that exposing the verifier directly to SQL structure provides the most effective signal for semantic verification.

\begin{table}[t]
\centering
\caption{Impact of ORM model size with the generator fixed to OmniSQL-7B. Execution Accuracy (EX, \%) and Pass@32 are reported on \textit{BIRD dev}, \textit{Spider dev}, and \textit{Spider test}. Best results per metric and dataset are shown in \textbf{bold}.}
\label{tab:orm_size_impact}

\resizebox{0.95\columnwidth}{!}{%
    \begin{tabular}{l
                    S[table-format=2.2]
                    S[table-format=2.2]
                    S[table-format=2.2]
                    S[table-format=2.2]
                    S[table-format=2.2]
                    S[table-format=2.2]}
    \toprule
    \multirow{2}{*}{\textbf{ORM}} &
    \multicolumn{2}{c}{\textbf{BIRD dev}} &
    \multicolumn{2}{c}{\textbf{Spider dev}} &
    \multicolumn{2}{c}{\textbf{Spider test}} \\
    \cmidrule(lr){2-3}\cmidrule(lr){4-5}\cmidrule(lr){6-7}
    & \textbf{EX} & \textbf{Pass@32} & \textbf{EX} & \textbf{Pass@32} & \textbf{EX} & \textbf{Pass@32} \\
    \midrule
    OmniSQL-7B  & 68.90 & 80.57 & 84.53 & 91.68 & 87.47 & 93.29 \\
    OmniSQL-14B & \textbf{69.23} & 80.57 & \textbf{85.01} & 91.68 & \textbf{87.84} & 93.29 \\
    OmniSQL-32B & 68.19 & 80.57 & 84.23 & 91.68 & 86.91 & 93.29 \\
    \bottomrule
    \end{tabular}
}
\end{table}

\paragraph{Scaling ORMs Beyond 7B}
Table~\ref{tab:orm_size_impact} shows that increasing ORM size beyond 7B parameters yields only marginal and inconsistent improvements. While a 14B ORM occasionally improves over 7B, a 32B ORM often underperforms, indicating diminishing returns. Overall, lightweight ORMs achieve a favorable balance between performance and efficiency, and larger verifiers cannot compensate for generator limitations.

\begin{table}[t]
\centering
\caption{Ablation study comparing autoregressive and Binary Cross-Entropy (BCE) loss fine-tuning on \textit{BIRD dev} across query difficulty levels. Both the generator and ORM model are OmniSQL-7B. Reported metric is Execution Accuracy (\%). Best results are in \textbf{bold}.}
\label{tab:finetune_ablation}
\sisetup{
  table-format=2.2,
  output-decimal-marker = {,} 
}
\renewcommand{\arraystretch}{1.08}

\resizebox{0.95\columnwidth}{!}{%
    \begin{tabular}{l S S S S}
    \toprule
    \textbf{Method} & \textbf{Simple} & \textbf{Moderate} & \textbf{Challenging} & \textbf{Total} \\
    \midrule
    Majority Voting & 72.86 & 59.70 & 52.41 & 66.95 \\
    BCE FT   & 72.86 & 59.91 & 51.72 & 66.95 \\
    Autoregressive FT    & \textbf{74.15} & \textbf{62.93} & \textbf{54.48} & \textbf{68.90} \\
    \bottomrule
    \end{tabular}
}
\end{table}

\paragraph{Fine-Tuning Objective}
Table~\ref{tab:finetune_ablation} compares autoregressive fine-tuning with Binary Cross-Entropy (BCE). Autoregressive training consistently outperforms BCE and Majority Voting, with the largest gains observed on moderate and challenging queries. These results indicate that modeling verification as an autoregressive generation task better captures semantic correctness than direct classification.

\paragraph{Summary}
Across all ablations, ORMs demonstrate stable behavior under varying candidate budgets, generator scales, prompts, and training objectives. Performance is primarily driven by the quality of feedback and verification signal rather than verifier size, reinforcing the effectiveness of lightweight, autoregressive ORMs for test-time verification.


\section{Conclusion}\label{sec:conclusion}
In this work, we studied the use of Outcome Reward Models (ORMs) as learned verification mechanisms for test-time inference in Text-to-SQL. Rather than introducing new reward modeling techniques, our goal was to investigate how existing ORM paradigms can be effectively adapted to structured query generation, where semantic correctness is critical.
We introduced \textit{GradeSQL}, a scalable pipeline for training task-specific ORMs through automated candidate generation and execution-based labeling, enabling verifier training without manual annotation and supporting deployment in standard Best-of-$N$ inference pipelines.
Across the BIRD and Spider benchmarks, ORM-based selection consistently improves execution accuracy over heuristic baselines such as execution-based Best-of-$N$ and Majority Voting. While absolute gains are moderate, they are stable across models and datasets, and become more pronounced in challenging queries and larger candidate regimes, where heuristic methods tend to saturate.
Overall, our findings suggest that learned verification provides a simple and effective complement to heuristic test-time selection strategies in Text-to-SQL, highlighting the importance of data curation and task-specific training for scalable verifier-based inference in structured reasoning tasks.
We release our code, datasets, and trained models to support reproducibility.

\section*{Limitations}
ORM-based selection requires an additional offline training phase, unlike heuristic baselines such as Majority Voting and execution-based Best-of-$N$. While this cost is amortized across queries, our comparison is not fully compute-matched end-to-end. Moreover, the observed gains, although consistent, are moderate (2--5\%) and mainly arise in challenging queries and larger candidate regimes, indicating improvements in selection rather than in the underlying generation capability.
Our approach relies on execution equivalence as a proxy for semantic correctness, which may fail to distinguish semantically equivalent queries or may reward spurious matches. In addition, evaluation is limited to clean benchmark datasets (Spider and BIRD), and performance in real-world settings with noisy schemas or ambiguous queries remains unclear. Finally, we focus on binary Outcome Reward Models; richer verification signals, such as fine-grained or process-level feedback, are not explored.

\section*{Ethical Considerations}
This work focuses on improving the reliability of LLM-based systems for structured data access through Text-to-SQL, with potential benefits for accessibility and decision support. However, incorrect or misleading SQL generation may lead to faulty data retrieval and downstream decisions, particularly in high-stakes domains. While ORM-based verification improves selection quality, it relies on execution-based supervision, which may inherit biases or errors present in underlying data or schemas. Our approach does not introduce new personal data or sensitive information, but care should be taken when deploying such systems on proprietary or privacy-sensitive databases. Finally, increased use of test-time compute and model-based verification may have environmental and cost implications. We encourage future work on robust evaluation in real-world settings and on developing more transparent and accountable verification mechanisms.


\bibliography{custom}

\end{document}